\pdfoutput=1

\newcommand\blfootnote[1]{%
  \begingroup
  \renewcommand\thefootnote{}\footnote{#1}%
  \addtocounter{footnote}{-1}%
  \endgroup
}

\documentclass[11pt]{article}

\usepackage{amsfonts,amsthm,bm}
\usepackage{subcaption}
\usepackage{graphicx}
\usepackage[]{EMNLP2022}

\usepackage{times}
\usepackage{latexsym}

\usepackage[T1]{fontenc}

\usepackage[utf8]{inputenc}

\usepackage{microtype}

\usepackage{inconsolata}
\usepackage{amssymb}
\usepackage{pifont}
\newcommand{\cmark}{\ding{51}}%
\newcommand{\xmark}{\ding{55}}%

\usepackage{amsfonts,amsthm,bm,amsmath}
\title{Beyond Model Interpretability: On the Faithfulness and Adversarial Robustness of Contrastive Textual Explanations}

\author{Julia El Zini \and Mariette Awad \\
Electrical and Computer Engineering Department \\
  American University of Beirut \\
  Beirut, Lebanon \\
  \texttt{jwe04@mail.aub.edu} \texttt{mariette.awad@aub.edu.lb}
}

\begin{document}

\maketitle

\begin{abstract}
\blfootnote{Papers accepted at Findings of Empirical Methods in Natural Language Processing (EMNLP), 2022.}

Contrastive explanation methods go beyond transparency and address the contrastive aspect of explanations. Such explanations are emerging as an attractive option to provide actionable change to scenarios adversely impacted by classifiers' decisions. However, their extension to textual data is under-explored and there is little investigation on their vulnerabilities and limitations. 

This work motivates textual counterfactuals by laying the ground for a novel evaluation scheme inspired by the \textit{faithfulness} of explanations. Accordingly, we extend the computation of three metrics, \textit{proximity}, \textit{connectedness} and \textit{stability}, to textual data and we benchmark two successful contrastive methods, POLYJUICE and MiCE, on our suggested metrics. Experiments on sentiment analysis data show that the connectedness of counterfactuals to their original counterparts is not obvious in both models. More interestingly, the generated contrastive texts are more attainable with POLYJUICE which highlights the significance of latent representations in counterfactual search. Finally, we perform the first semantic adversarial attack on textual recourse methods. The results demonstrate the robustness of POLYJUICE and the role that latent input representations play in robustness and reliability. 
\end{abstract}



\maketitle

\section{Introduction}
With the unprecedented growing deployment of Machine Learning (ML) models in high-stake areas such as law enforcement and medicine, many concerns are raised about their black-box decision-making process.
Recently, transparency is becoming a momentous requirement for accountable ML bringing forth the concept of Explainable AI (ExAI). ExAI is witnessing endeavors in all modalities and textual data in particular \cite{pruthi2020learning,ribeiro2016should,lundberg2017unified}. 
However, such explanations might not be sufficient in critical areas where stronger guarantees and more fine-grained explanations are required. Data controllers and subjects pose strong requirements on the \textit{usefulness} aspect of explanations which implies a selective, contrastive and social process \cite{are_explanations_useful,ribera2019can}. The latter entails an interaction between the explainers and the explainees and human-understandable explanations \cite{mittelstadt2019explaining}.

To this end, contrastive\footnote{Throughout this work, we use the terms \textit{contrastive}, \textit{counterfactual} and \textit{recourse} interchangeably.} explanations have seen a surge of interest as a main tool to reach recourse \cite{ustun2019actionable,mothilal2020explaining,madaan2021generate}. Those explanations search for optimally proximate alternative inputs that would result in a different, \textit{usually desired}, prediction. They offer explanations that are tailored to the recipient's beliefs and comprehension capabilities. 

Whilst there is a plethora of literature published on recourse methods applied on tabular datasets and computer vision applications \cite{mothilal2020explaining,dosovitskiy2016generating,pawelczyk2021carla}, little is available on contrastive textual explanations. Specifically, hand-crafted contrastive sets have been employed before to evaluate fairness and robustness of ML models \cite{garg2019counterfactual} by rewriting input instances \cite{gardner-etal-2020-evaluating} and defining perturbation functions \cite{ribeiro2020beyond} to obtain counterfactual sets. A novel, yet interesting, vein of research is considering an automated counterfactuals generation in NLP \cite{ross2021explaining,wu2021polyjuice}. The focus of our work is this largely under-explored \textit{targeted} recourse methods for language data and their evaluation schemes. We highlight scenarios where non-recourse methods fall short of the usefulness aspect of explanation especially due to the blend of syntax and semantics in the words. 


Additionally, we consider the assessment schemes, and we target a novel evaluation aspect of the plausibility and attainability aspect of the generated counterfactuals. We argue that counterfactuals should (1) meet textual attainability from a grammatical and semantic perspective, (2) convey connectedness to their original counterparts, and (3) satisfy local algorithmic stability. Accordingly, we extend \textit{proximity}, \textit{connectedness} and \textit{stability}, in the context of \textit{faithfulness}, to textual data and we propose tangible measures to quantify them. We benchmark our metrics on a sentiment analysis task on two famous recourse methods \cite{wu2021polyjuice,ross2021explaining}. Our results highlight the role that latent representations in \cite{wu2021polyjuice} play in robustness and plausibility. Finally, we present the first study on the resilience of textual recourse methods in the context of adversarial attacks. The study demonstrates a significant improvement in the adversarial robustness of POLYJUICE over MiCE. Our contribution falls under the following categories:
\begin{itemize}
    \item Surveying textual counterfactuals and highlighting their usefulness over traditional ExAI
    \item Proposing new evaluation metrics inspired by explanation \textit{faithfulness} and benchmarking contrastive methods, POLYJUICE and MiCE
    \item Evaluating the robustness of NLP recourse methods through semantic adversarial attacks 
\end{itemize}
Next, we start by motivating the use of textual counterfactuals in Section~\ref{sec:use_cases} and highlighting their interconnection to adversarial attacks in Section~\ref{sec:adversarial}. Then, we present the background needed on counterfactuals methods in Section~\ref{sec:survey_text} before reporting current evaluation schemes in Section~\ref{sec:eval_quant}. Finally, we extend the \textit{faithfulness} concept to textual data and validate it in Sections~\ref{sec:faithfulness} and \ref{sec:validation} respectively before concluding with final remarks in Section~\ref{sec:conc}. 

\section{Use cases of Contrastive Explanations}\label{sec:use_cases}
\subsection{Favourable Use-cases}
Textual ExAI methods interpret outcomes by highlighting segments that support the decision \cite{ribeiro2016should,ribeiro2018anchors} by computing gradients, attention weights, and simpler approximations. Such methods have fundamental impediments. 

\begin{figure}
    \centering
    \includegraphics[width=0.45\textwidth]{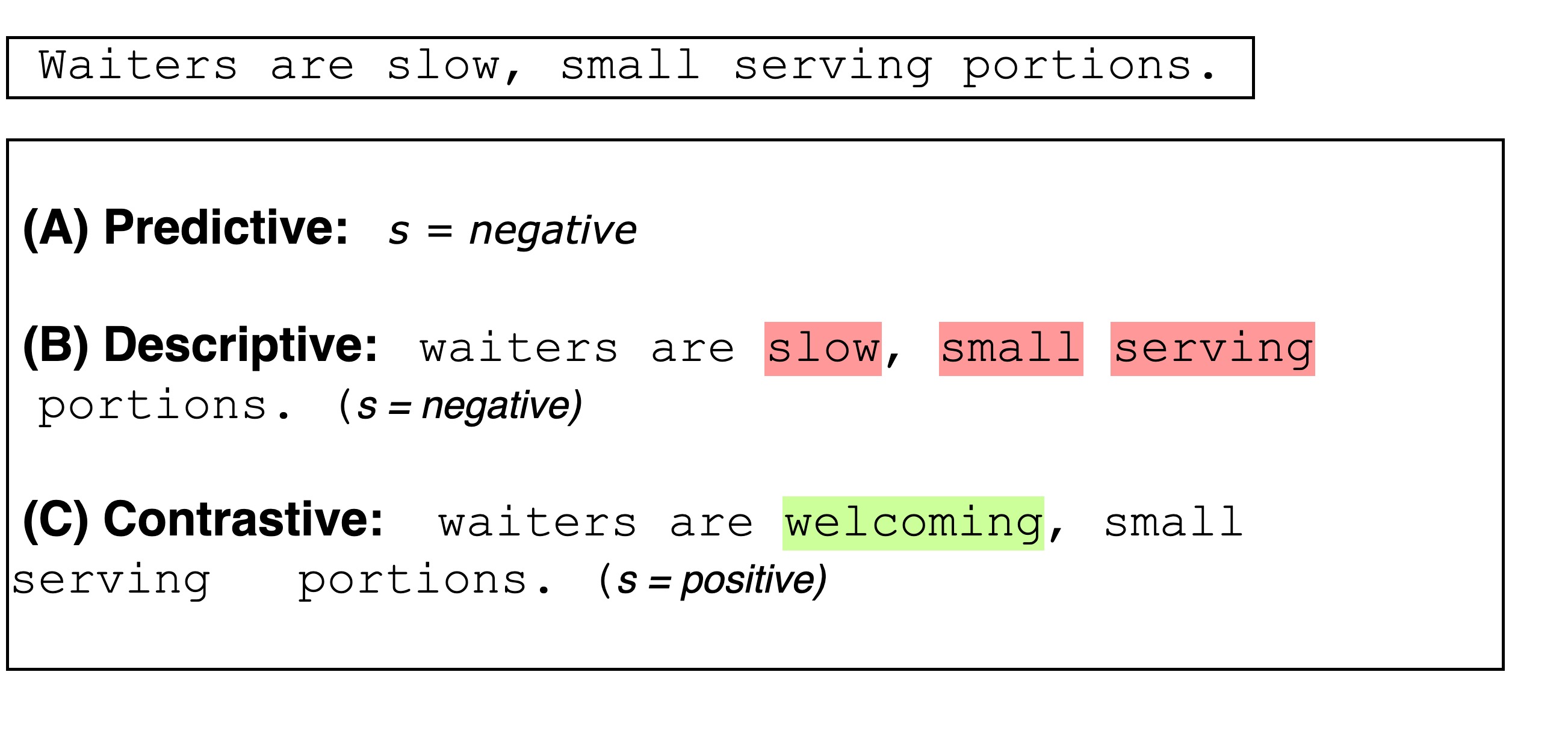}
    \caption{Non-recourse Vs. recourse methods}
    \label{fig:reviews_case_study}
\end{figure}

First, highlighting important input segments fails to specify the contrast between different decision boundaries. 
Second, the search space of traditional explainability is restricted to the words in the input text. Such methods forsake an integral space of words whose absence from inputs affected the prediction. This assumption thwarts the comprehensiveness and completeness of the explanations.
Finally, the fusion of syntax and semantics in a word makes it hard for a practitioner to identify the precise aspect of the word the model is attending to. For instance, the explanation in Figure~\ref{fig:reviews_case_study} shows that the sentiment is negative because of the word \textit{slow}. A user is left uninformed on whether the sentence structure (part-of-speech tag, named entity...) or the meaning of the adjective ``slow'' is driving the model's decision.  
    
While the first two limitations are shared by general explainability methods; the last limitation is specific to NLP.  Recourse methods address these limitations and present additional assets to the model's transparency. Their underlying design matches the human perception of explanations. In fact, humans inherently understand explanations contrastively \cite{kumar2020problems}. A fact-foil contrast can thus introduce a user-centered explanation aspect that complies with the human-in-the-loop drift in AI. 

To further highlight the importance of counterfactuals in NLP, we show how they can be leveraged in digital strategy \cite{boulton_writer_2019}. We assume an ML model $\mathcal{M}$ that classifies customer comments based on sentiment. We consider three scenarios thereafter illustrated in Figure~\ref{fig:reviews_case_study}.

\textbf{(A) Predictive:} $\mathcal{M}$ predicts a sentiment $s\in S = \{$ positive, negative, neutral $\}$.

\textbf{(B) Descriptive:} $\mathcal{M}$ predicts a sentiment $s\in S$, with a set of input segments supporting the decision, i.e. through non-recourse explainable AI. 

\textbf{(C) Prescriptive:} $\mathcal{M}$ predicts a sentiment $s\in S$, with one (or more) counterfactual inputs that are close to the original input but can change the decision $s$. 

\textbf{(A)} helps the institution in the assessment of their customer satisfaction. A company that hopes to better understand its customers has to resort to \textbf{(B)} or \textbf{(C)}. Both models analyze user feedback as detractors to identify areas that need improvement. \textbf{(B)} highlights the service and the portion size. However, these explanations are not the strong guarantee that strategic planning requires. On the other hand, \textbf{(C)} does not stop at highlighting input features that support $\mathcal{M}$'s decision; it generates reviews in a parallel \textit{counterfactual} word that can change the feedback of the user from \textit{negative} to positive. \textbf{(C)} can thus prescribe a workable strategy that is more likely to improve the users' feedback.

In addition to that, contrastive methods convey a way to evaluate and improve models and expose bias \cite{ribera2019can,wu2021polyjuice,sharma2020certifai,cheng2020fairfil,garg2019counterfactual,huang2020reducing}. Contrast sets can also serve as ways to identify general model's vulnerability \cite{gardner-etal-2020-evaluating,artelt2021contrastive} and improve the robustness through contrastive data augmentation \cite{wu2021polyjuice,qi2020small}. 
Contrastive learning is also shown to promote better separation in clustering applications \cite{zhang2021supporting} and easier model debugging for non-experts \cite{ribera2019can}. \cite{iter2020pretraining} and \cite{kiyomaru2021contextualized} further prove that pre-training language models with contrastive text improves the discourse coherence between clauses in text. 

\subsection{\textit{Malignant} Counterfactuals}\label{sec:adversarial}
\textit{Can counterfactuals be used as adversarial attacks?} To illustrate this, we consider hate speech detection where counterfactuals can be used to make a post get through the ``hate speech'' check. These modification are \textit{minimal} by design. This brings us to jointly study contrastive explanations (CEs) through the lens of adversarial attacks (AEs) and robustness. 

While both methods solve a similar optimization problem, the philosophy behind the optimization of CE and AEs can be conflicting. The former methods compute alternatives that result in a different \textit{desired} prediction.  Adversarial alternatives are generally semantically indistinguishable from the original input \cite{zhang-etal-2019-generating-fluent} whereas counterfactuals are perturbations to actionable features. Interested readers are referred to the work of \cite{pawelczyk2021connections} that establishes the theoretical and empirical connections between the literature on counterfactual explanations and adversarial examples.

\section{Contrastive Textual Explanations}\label{sec:survey_text}
\subsection{General Contrastive Theory}\label{sec:general_contrastive}
\textit{Assuming a predictor, potentially non-linear, $f: \mathcal{X} \mapsto \mathcal{Y}$, an instance $\bm{x^i} \in \mathcal{X}$ such that $f(\bm{x^i}) = y_{\text{fact}}$ and a foil class $y_{\text{foil}}$, a counterfactual $\bm{x}_{cf}^i \in \mathcal{X}$ can be computed as:}
\begin{align}\label{eq:formulation}
    \underset{\bm{x}_{cf}}{\arg \min} &\hspace{1em} d(\bm{x}_{cf}^i , \bm{x^i})\\
    \text{subject to} &\hspace{1em} f(\bm{x}_{fl}^i) =  y_{\text{foil}}
\end{align}
where $d(.)$ is a distance metric.

This optimization can be also perceived as $\underset{\bm{x}_{cf}}{\arg \min} \hspace{1em} L(f(\bm{x}_{fl}^i), y_{\text{foil}}) + \lambda d(\bm{x}_{cf}^i , \bm{x^i})$  in the Lagrangian notation, with $l(.)$ denoting a loss function and $\lambda>0$ is a regularization factor that balances the minimal edit distance and the success in altering the model's decision. Depending on the method, restrictions might complement the above definition. 

One line of research deploys gradient-based techniques to achieve a feasible solution \cite{dhurandhar2018explanations,schut2021generating}. Another track focuses on graph search techniques \cite{poyiadzi2020face}, growing hyper-spheres \cite{laugel2017inverse} and integer programming tools \cite{ustun2019actionable}. 

When considering the counterfactual cost or distance $d(.)$, the literature has formed a consensus on the use of the (normalized) $l_0$ or $l_1$ norm or any convex combination thereof \cite{mothilal2020explaining,karimi2020model,ustun2019actionable,wachter2017counterfactual}. 
ExAI is witnessing momentum in the adoption of manifold-like distance measures based on adversarial learning \cite{zhou2021cdl,dosovitskiy2016generating} and Variational Auto-Encoders (VAEs) \cite{samanta2020vae,joshi2019towards}. Interested readers are referred to \cite{survey} and \cite{karimi2020survey} that survey state-of-the-art recourse methods. 
\subsection{Contrastive Methods in NLP}
Up until 2021, natural language data was a modality that did not receive enough attention when it comes to recourse methods. Although contrastive sets have been extensively applied to evaluate robustness and fairness \cite{garg2019counterfactual,huang2020reducing,gardner-etal-2020-evaluating} in NLP, these sets are carefully manufactured as adversarial attacks. 

\cite{jacovi2021contrastive} present contrastive explanations as textual highlights that support one decision against its contrasts. Their approach encodes an input text into a latent vector and applies a linear interpolation where the contrastive direction is found in the latent space. 
This approach can be viewed as a contrastive search based on the positive pertinent which is not very aligned with the ``actionable change'' of recourse methods. \cite{dhurandhar2018explanations} formulate the pertinent negatives aspect of counterfactuals that are represented by the \textit{missing} features but did not apply their pertinent negatives to natural language data.

Generate Your Counterfactuals (GYC) in \cite{madaan2021generate} is one of the first explicit attempts at automating the generation of textual contrastive explanations. GYC trains a language model to reconstruct the input from the generated counterfactuals and then learns perturbations on the latent space while forcing a proximity constraint. 

\cite{ross2021explaining} refer to the GYC's proximity criterion as \textit{minimal edits} in their \textbf{Mi}nimal \textbf{C}ontastive \textbf{E}dits framework. To this end, they train a \textit{contextualized} EDITOR to associate edits with task-specific labels by applying masks on input segments that are important for a particular label. EDITOR then serves as a generator by predicting the labels of some masked inputs monitored by binary and beam search to find the optimal maskings.

POLYJUICE \cite{wu2021polyjuice} allows for more edits through negation, replacement, insertion, and deletion for targeted counterfactuals monitored by control codes. Similar to GYC, POLYJUICE relies on a language model to achieve a \textit{fluent} conditional text generation. A filtering layer is added to the process to refine the generated counterfactuals by ignoring the ones that achieve low fluency scores. 

The Contrastive Attributed explanations for Text (CAT) of \cite{chemmengath2021let} inject an attribute prediction layer in the contrastive search process. This layer indicates attributes that the contrast adds to or removes from the given example. 
%
Very recently, Malandri et al. \cite{malandri2022contrxt} develop ContrXT, a \textbf{T}ime \textbf{Contr}astive model-agnostic e\textbf{x}planation framework in lifelong learning settings. ContrXT is not restricted to locally explaining predictions, it rather focuses on the learning process and on how the decision paths of classifiers evolve after retraining. 
The discussed methods are summarized in Table~\ref{tbl:survey}.

\begin{table*}[h]
\small
\centering
\begin{tabular}{p{3cm}p{1.8cm}p{1.8cm}p{1.8cm}p{5.6cm}}

Method                                                   & Pertinent Negatives    & Diversity             & Latent Representations & Strategy                                                                     \\
\hline \hline
Highlights & \xmark & \xmark & \cmark  & Linear interpolation on latent space                                         \\
GYC   & \cmark  & \cmark & \cmark  & Perturbations on latent space and auto-encoder generation \\
MiCE    & \cmark  & \cmark & \xmark  & Masking input segments and searching for optimal mask combinations           \\
POLYJUICE  & \cmark  & \cmark & \cmark  & Conditional generation with refinement                                  \\
CAT  & \cmark  & \xmark & \xmark  & Attribute injection in the contrastive search process                        \\
Contr-XT                                                 & \cmark  & \xmark & \xmark  & Global and time-sensitive explanations through BDD  \\
\hline
\end{tabular}
\caption{Summary of existing work on contrastive textual explanations}\label{tbl:survey}
\end{table*}

\section{Evaluation Methods}\label{sec:eval_quant}


\subsection{Quantitative Evaluation}
\sloppy The most intuitive desiderata for any contrastive explanation are their proximity and ability to change the model's prediction. Both conditions are axiomatically inferred from the problem formulation in Equation~\ref{eq:formulation}. In NLP, these conditions are referred to as \textit{minimal distance} and label-flip score respectively. Proximity between $\bm{x}_{cf}$ and $\bm{x}$ is measured by word-level Levenshtein distance \cite{levenshtein1966binary} reflecting the edit distance in terms of replacement, insertions and deletions. 
We draw the reader's attention to the fact that embedding distance measures how similar two vectors are in terms of syntax and semantics \cite{vylomova2016take} whereas Levenshtein distance reflects the edit distance, or the path to reach counterfactuals. The latter is aligned with the fundamentals of contrastive textual explanations whereas the former is used to measure content preservation. Another way to measure the edit distance is through syntactic trees \cite{zhang1989simple,wu2021polyjuice}.

An additional requirement for counterfactuals is the diversity of the generated explanations. Inspired by the Self-BLEU metric of \cite{zhu2018texygen}, diversity can be measured through the Self-BLEU or Self-BERT \cite{zhang2019bertscore} metric between the generated counterfactual samples. 

Other requirements that are tailored to natural language are (1) fluency through grammatical correctness and semantic meaningfulness, and (2) content preservation. 
Fluency can be evaluated by comparing the loss of a particular language model on $\bm{x}_{cf}$ and $\bm{x}$ using a pre-trained model \cite{ross2021explaining,morris2020textattack,wu2021polyjuice}.
Content preservation can be inferred by latent embedding representations as the cosine similarity between the embeddings of $\bm{x}_{cf}$ and $\bm{x}$.

\subsection{Qualitative Evaluation}\label{sec:eval_qual}
In the social aspects of AI, user studies are ubiquitous in evaluating explainable and fair AI models \cite{luss2021leveraging,natesan2020model,singh2018hierarchical}. GYC uses a score to estimate the human judgment of grammatical correctness, plausibility, fluency, sentiment change, and content preservation. Similarly, CAT evaluates human judgment of completeness, sufficiency, satisfaction, and understandability mainly. Instead of surveying human judgment, MiCE's counterfactuals are compared to human edits for overlap, minimality, and fluency. Finally, ContrXT employs crowd-sourcing efforts to evaluate its global explanations, their understandability, and usefulness.

\section{Faithfulness Metrics }\label{sec:faithfulness}
None of the metrics discussed so far explicitly targets explanation faithfulness that has been studied in non-textual frameworks \cite{laugel2019issues,pawelczyk2020learning}. In this work, we redefine \textit{faithfulness} \cite{laugel2019issues}, in natural language settings, via three main requirements.

Explanation \textit{faithfulness} entails that counterfactuals should be generated from a ``possible'' world which is proximate to the starting point specified by the user. This is formalized in two quantitative measures: proximity and connectedness \cite{laugel2019issues} inferring an \textit{attainable} generation of counterfactuals based on the input distribution. Moreover, faithfulness to the explainee engenders local stability of the explainer during the counterfactual generation process. 


\subsection{Proximity} 
The contrastive explanation is only useful when presented in terms of plausible means of action. A plausible contrastive text is usually proximate (using a distance notion) to a ground-truth text from the same foil class. 

Formally, we consider an instance $\bm{x}$ belonging to the fact class $y_{\text{fact}}$ and its counterfactual $\bm{x}_{cf}$ belonging to the foil class $y_{\text{foil}}$. Proximity of $\bm{x}_{cf}$ is measured as the ratio between its distance to $\bm{x}$ and the minimum distance between $\bm{x}$ and a ground truth input belonging to the foil class, $\mathcal{X}^{\text{foil}}$. 
\begin{equation}\label{eq:proximity}
    P(\bm{x}_{cf}) = \frac{d(\bm{x}, \bm{x}_{cf})}{\underset{\bm{x}_{gt} \in \mathcal{X}^{\text{foil}}}{\min}d(\bm{x}, \bm{x}_{gt})}
\end{equation}
The notion of proximity is not any different in NLP, except for the computation of the distance metric. This will be discussed later in this section.

Furthermore, we propose the Local Reachability Density (LRD) as a quantitative measure for proximity. LRD reflects how far a point ($\bm{x_{cf}}$) is from the nearest cluster of points ($\bm{x}_{gt} \in \mathcal{X}^{\text{foil}}$). Mainly, 

\begin{equation}
    LRD_k(\bm{x}_{cf}) = 
    \frac{1}{\sum_{\bm{x}_{gt} \in \mathcal{N}_k(\bm{x}_{cf}) \cap \mathcal{X}^{\text{foil}}}
    \frac{RD(\bm{x}_{cf},\bm{x}_{gt})}{||\mathcal{N}_k(\bm{x}_{cf}) ||}}
\end{equation}
with RD is the reachability distance defined as $RD(i,j) = \max \big(k-\text{distance}(i), d(i,j)\big)$, with $k-$distance$(i)$ is the distance from $i$ to its closest neighbor. Higher LRD values are desired as they reflect closer clusters of ground truth data. 

\subsection{Connectedness}
Counterfactuals plausibility also engenders a continuous connectedness to the original ground-truth observation. Relying on the topological notion of the path, we borrow the definition of connectedness from \cite{laugel2019issues} as follows:

($\epsilon-$connectedness) \textit{$x_1 \in \mathcal{X}$ is $\epsilon-$connected to $x_2 \in \mathcal{X}$ if $f(x_1) = f(x_2)$ and $\exists$ an $\epsilon$-chain $(e_i)_{i<N}\in \mathcal{X}^N$ between $x_1$ and $x_2$ such that, $e_0 = X_1$, $e_N = x_2$ and $\forall i < N d(e_{i}, e_{i+1}) < \epsilon$ $\forall n<N, f(e_i) = f(e).$}

\sloppy The implementation of the above definition is a compelling problem. We highlight its analogy with density based clustering \cite{ester1996density,laugel2019issues} and we adopt the use of DBSCAN algorithm to check whether two texts are connected while setting the $min\_points$ parameter to 2. 

\subsection{Stability}
This criterion is highly related to robustness where stability requires close counterfactuals for close inputs. Formally, a contrastive explanation method is stable if it satisfies
    \begin{equation}\label{eq:stability}
        \underset{\bm{x'}\in\mathcal{B}(\bm{x}, \epsilon)}{\max} \frac{d(\bm{x}_{cf}', \bm{x_{cf}})}{d(\bm{x}, \bm{x'})} < \epsilon_2
    \end{equation}
    with $\mathcal{B}(x,r)$ is a ball of center $x$ and radius $r$.
    
Stability can also serve as an evaluation of the model resistance to adversarial attacks. \cite{slack2021counterfactual} bring to the front the high sensitivity of recourse methods to insignificant input changes. In fact, gradient-based methods \cite{wachter2017counterfactual} can yield counterfactuals $\bm{x}^{(1)}_{cf}$ and $\bm{x}^{(2)}_{cf}$ that are very distant for close inputs $\bm{x}^{(1)}$ and $\bm{x}^{(2)}$. 

In summary, \textit{proximity} measures the distance between the encodings of the generated counterfactual $x'$ and the closest instance with the same label in the ground truth data. Connectedness requires $x'$ to be accessible from $x$ along a path consisting of neighboring points with the same label. \textit{Stability} requires the close counterfactuals for close inputs.

\textbf{Distance measure:}
The aforementioned notions anticipate a distance measure. Choosing a suitable distance measure is of utmost criticality, especially in NLP, where distances have to reflect syntax and semantics. We use latent space embeddings encoded by language models such as GPT-2 \cite{radford2019language} and we compute their cosine similarity.

\subsection{Discussion}
Inadvertently, some of the metrics discussed in Section~\ref{sec:eval_quant} hint to the faithfulness concept. In general, contrastive attainability can be linked to fluency and grammatical correctness \cite{wu2021polyjuice,chemmengath2021let,ross2021explaining}. However, none of the existing work on textual contrastive explanations explicitly addresses \textit{faithfulness}. 
We draw the reader's attention to the fact that \textit{proximity} is not to be mistaken with the minimal edit distance requirement. The former is the smallest distance between the counterfactual and a ground truth instance belonging to the same class and the latter is a minimal distance between the counterfactual and the original instance. 

The masking strategy in MiCE can be associated with a search strategy for the topological path. However, this aspect is not explicitly tested. The content preservation concept of \cite{chemmengath2021let} might be an educated guess at \textit{connectedness} but not a direct measure. 

\section{Validation of Faithfulness}\label{sec:validation}
We consider models with open source code, POLYJUICE, MiCE, and ContrXT mainly. Counterfactuals generated by ContrXT are global which makes \textit{faithfulness} not directly applicable as it evaluates specific (local) explanations. Thus, we consider POLYJUICE and MiCE for our validation. We train both models on the IMDB sentiment analysis task on NVIDIA K80/T4 GPU with 16GB RAM. We consider restaurant reviews for sentiment analysis \footnote{kaggle.com/apekshakom/sentiment-analysis-of-restaurant-reviews}  with 977 validation instances. 



\subsection{Proximity}
\begin{figure}[h]
\centering
\begin{subfigure}{0.25\textwidth}
	\centering
	\includegraphics[width=0.9\linewidth]{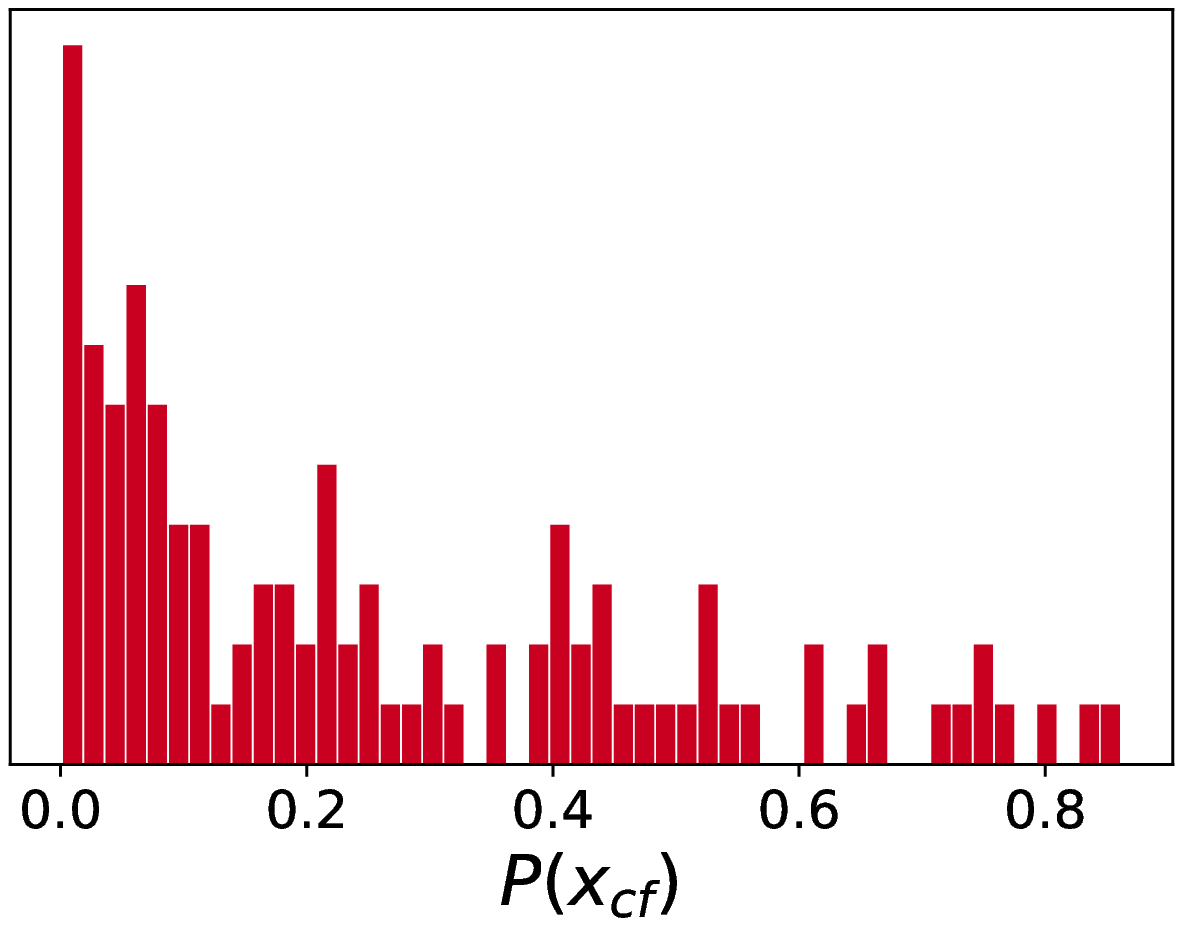}
	\caption{POLYJUICE}
\end{subfigure}%
\begin{subfigure}{0.25\textwidth}
  \centering
  \includegraphics[width=0.9\linewidth]{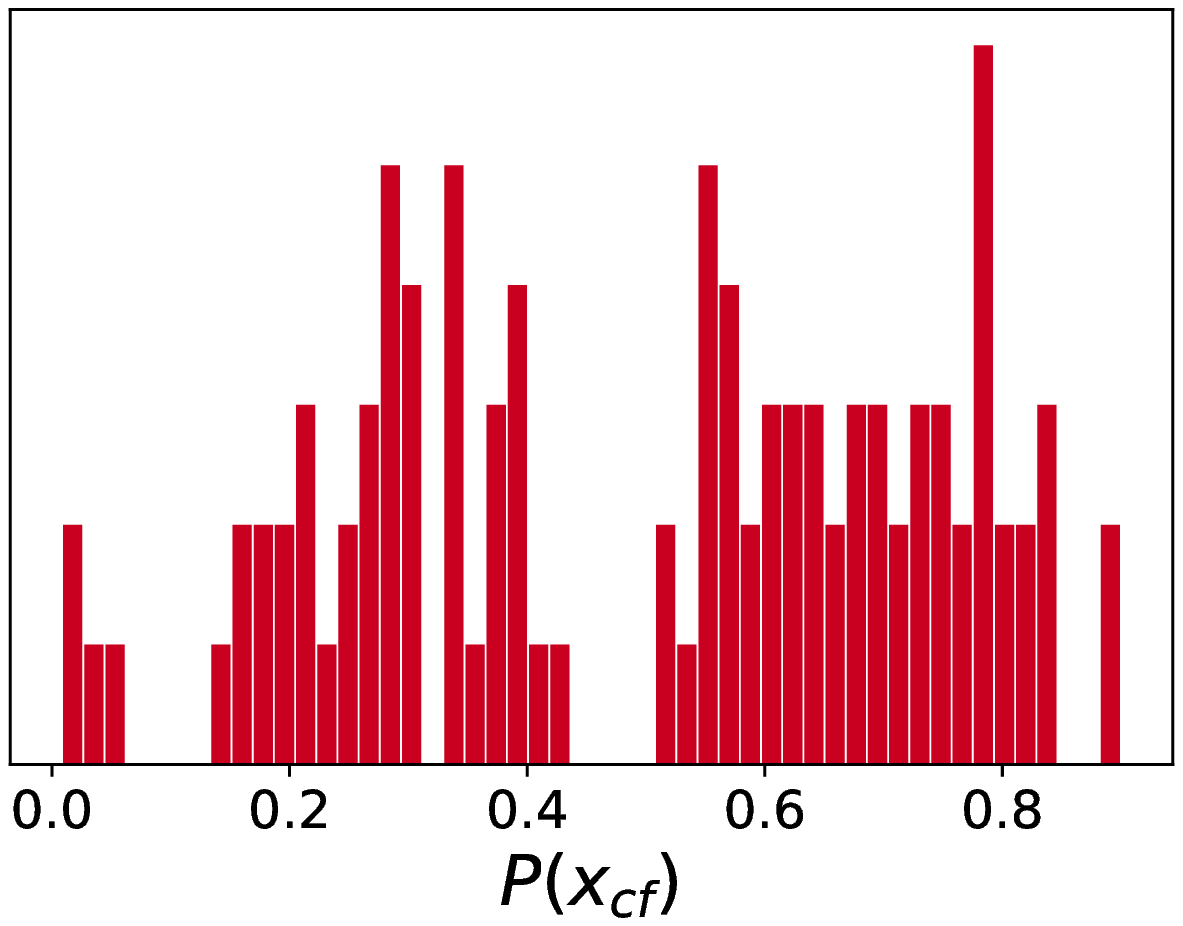}
  \caption{MiCE}
\end{subfigure}
\caption{Distribution of $P(\bm{x}_{cf})$ scores}
\label{fig:P_score_dist}
\end{figure}

We start by evaluating how close the generated counterfactuals are to ground truth data from the foil class. For this purpose, we compute $P(\bm{x}_{cf})$ of Equation~\ref{eq:proximity} and plot the distribution of its values in Figure~\ref{fig:P_score_dist}. One can see a predominance of low proximity scores ($<0.2$) in POLYJUICE and an inclination to achieve higher scores with MiCE. 

\begin{figure}[h]
\centering
    \begin{subfigure}{0.4\textwidth}
    \centering
    \includegraphics[width=\textwidth]{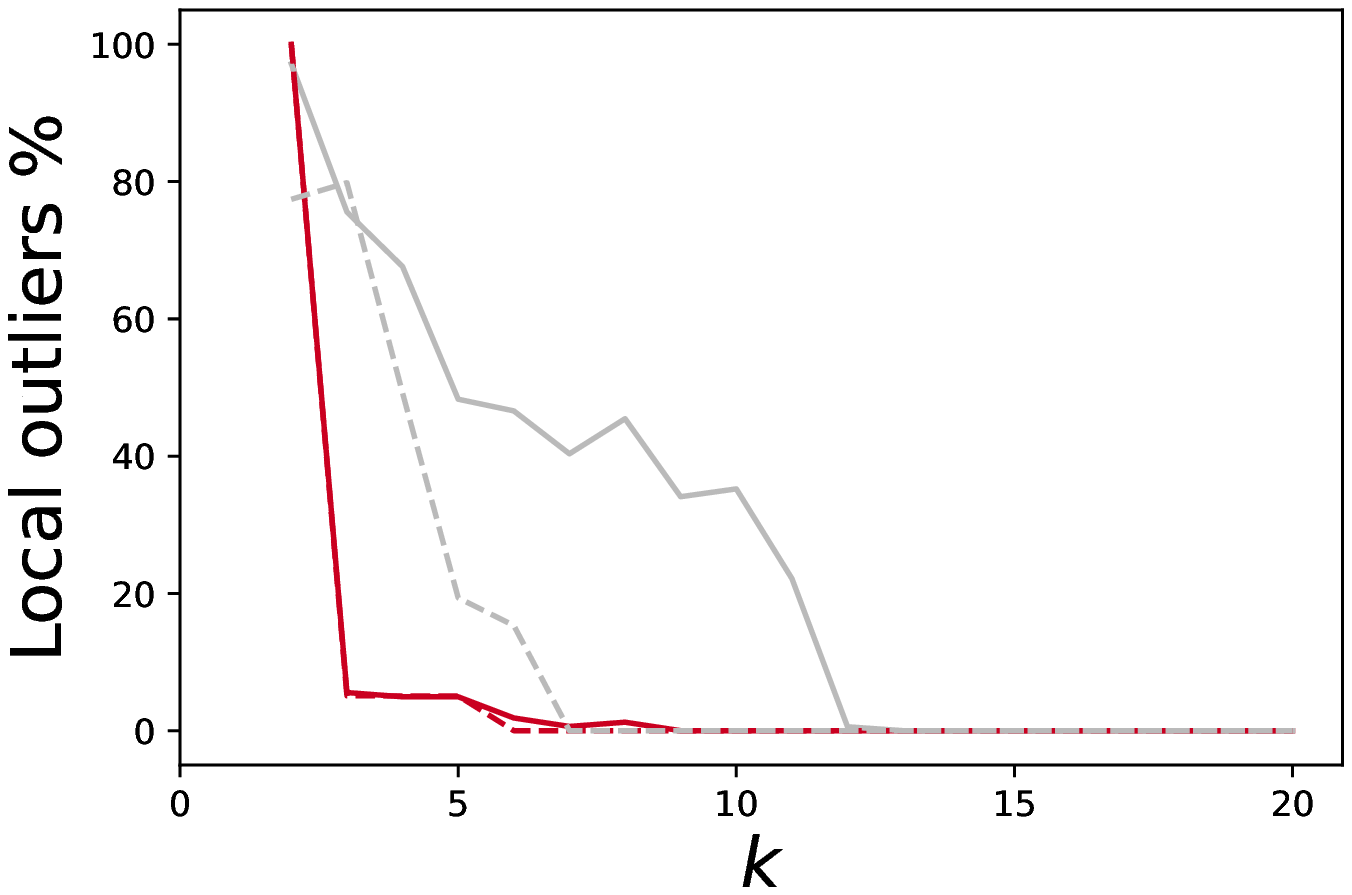}
    \caption{Proximity scores}
    \label{fig:proximity}
    \end{subfigure}
    \begin{subfigure}{0.4\textwidth}
    \centering
    \includegraphics[width=\textwidth]{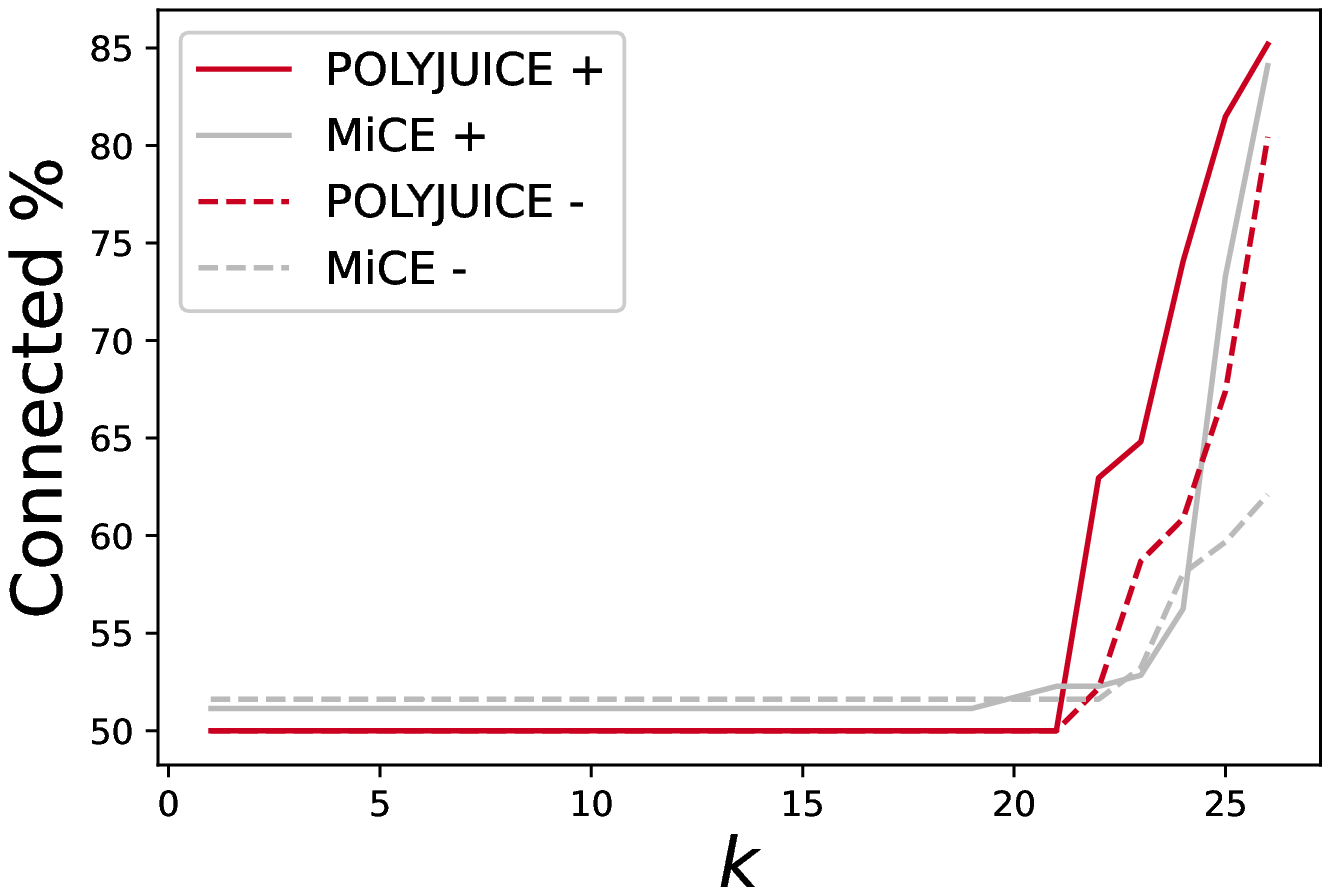}
    \caption{Connectedness scores}
    \label{fig:connectedness}
    \end{subfigure}
    \caption{Scores while changing the number of neighbors $k$}
\end{figure}

We further split our validation data according to their foil classes into two categories: positive and negative sentiment foils. For both categories, we compute the outlier factor for the generated counterfactuals, which is inversely proportional to LRD, while changing $k$ and we show the values in Figure~\ref{fig:proximity}. For small $k$, i.e. strong conditions on outliers, a great deal of the generated counterfactuals, especially with POLYJUICE, are considered outliers. With fair values of $k$, POLYJUICE drops its generated outliers to nearly zero while some outliers can still be observed with MiCE. Both explanation models are systematic with the foil class being positive or negative sentiments. 

\subsection{Connectedness}
To assess whether the generated counterfactuals are connected to their original factuals, we compute the connectedness score for both explanation models and foil sentiments. The results shown in Figure~\ref{fig:connectedness} demonstrate that POLYJUICE and MiCE achieve low connectedness scores when $k$ is small where only half of their generated counterfactuals can be considered connected to the original input. When we loosen the connectedness requirement by increasing $k$, we notice that more counterfactuals become connected especially with POLYJUICE. For both explanation methods, positive sentiment foil classes seem to achieve higher connectedness scores but the discrepancy between positive and negative sentiments is insignificant with MiCE.

\subsection{Stability}
\begin{figure}[h]
	\centering
	\begin{subfigure}{0.24\textwidth}
		\centering
		\includegraphics[width=0.9\linewidth]{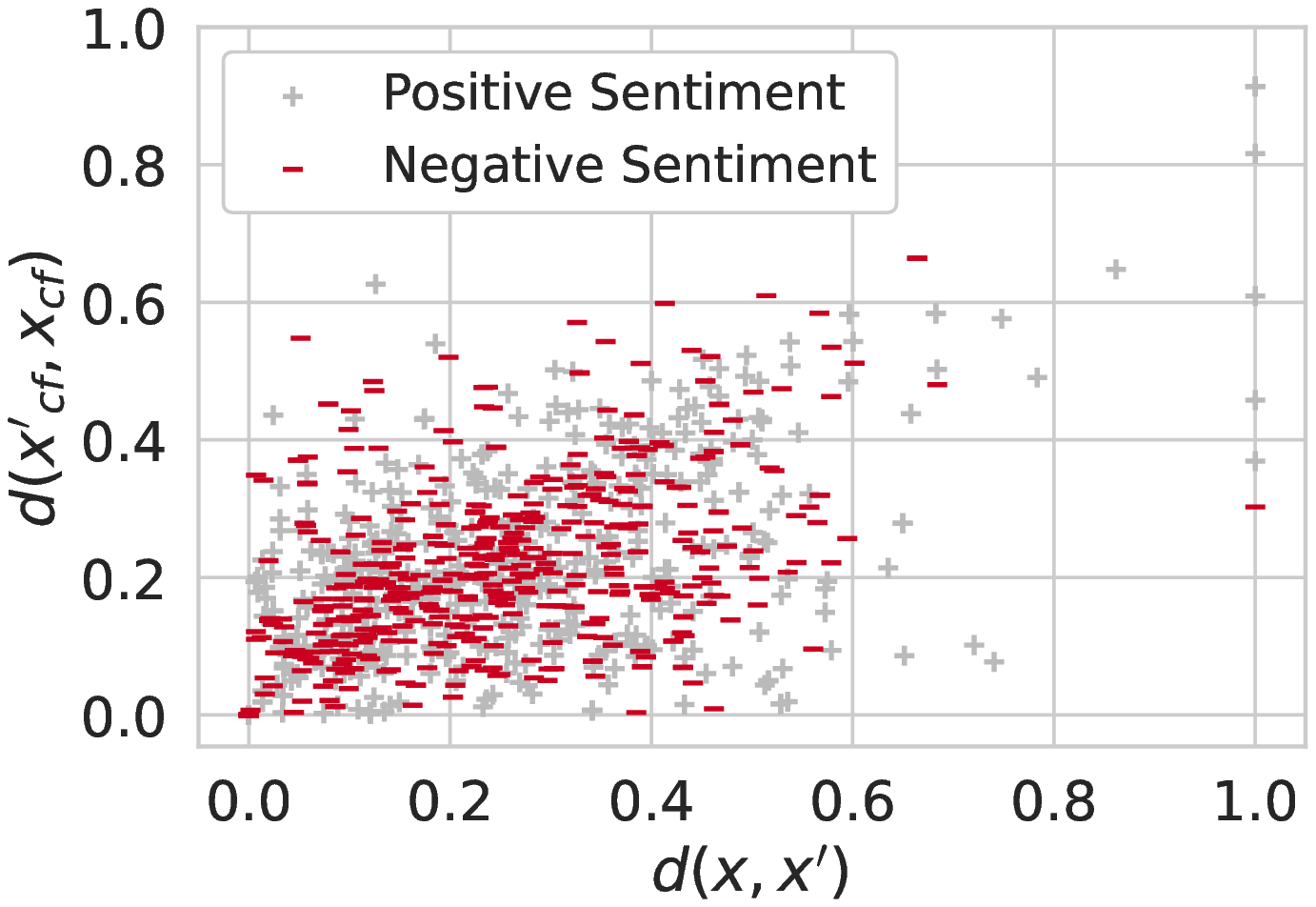}
		\caption{POLYJUICE}
	\end{subfigure}%
	\begin{subfigure}{0.24\textwidth}
		\centering
		\includegraphics[width=0.9\linewidth]{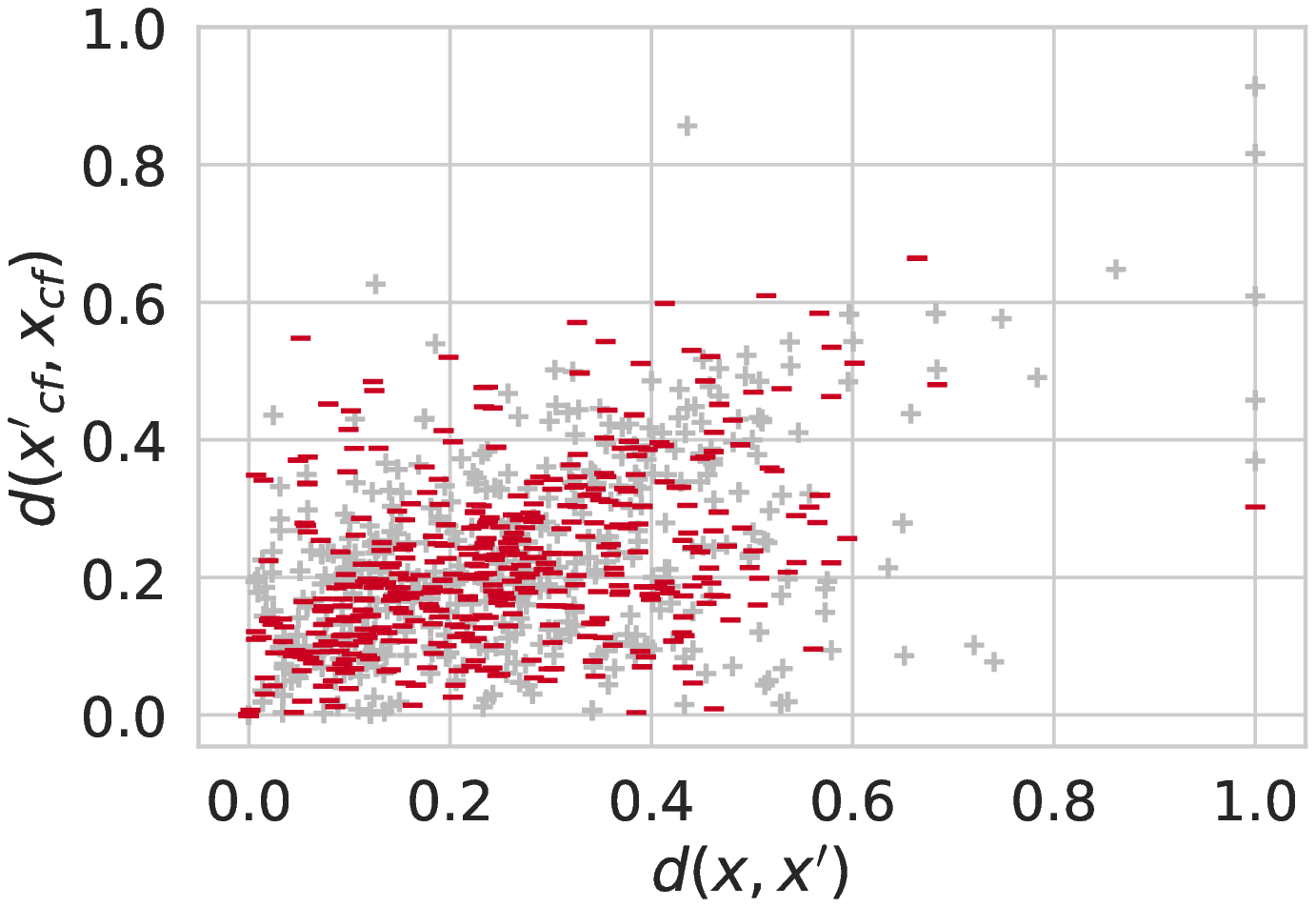}
		\caption{MiCE}
	\end{subfigure}
	\caption{Scattering of counterfactual similarity with respect to the input similarity. Linear scattering infers local stability. }
	\label{fig:stability_scatter}
\end{figure}

We compute $d(\bm{x}_{cf}', \bm{x_{cf}})$ as counterfactual similarity and $d(\bm{x}, \bm{x'})$ as input similarity and show how the former measure is scattered in terms of the latter in Figure~\ref{fig:stability_scatter} for POLYJUICE and MiCE. Both plots show that a near-linear correlation governs both models with some high variance. The ratio $\frac{d(\bm{x}_{cf}', \bm{x_{cf}})}{d(\bm{x}, \bm{x'})}$ represented by the slope of the linear regression model on the given scatter plots is bounded showing a stability of both explanation algorithms. This can suggest that the non-gradient aspect of the considered contrastive methods yields more robust counterfactuals.
The lower variance in POLYJUICE suggests better robustness guarantees. Besides, no significant distinction can be inferred between the two foil categories.

\begin{figure}[h]
    \centering
    \includegraphics[width=0.5\textwidth]{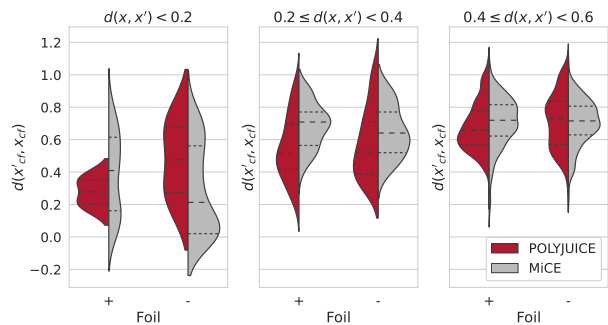}
    \caption{Distribution of the distance between counterfactuals for different input distance ranges}
    \label{fig:stability_violin}
\end{figure}

Finally, we consider more fine-grained stability study, by considering three ranges of input similarities: $d(x, x')<0.2$, $0.2\le d(x, x')<0.4$ and $0.4\le d(x, x')<0.6$. Figure~\ref{fig:stability_violin} shows how the counterfactual similarity is distributed for the three considered ranges.  Locally, i.e. with input distance $<0.2$ POLYJUICE is shown to be more stable on the positive foil class by achieving low distances on the generated counterfactual. MiCE seems to outperform POLYJUICE on the negative foil class. Zooming out, better stability is observed with POLYJUICE for both foil classes.

\subsection{Adversarial Robustness}
\begin{figure}[h]
	\centering
	\begin{subfigure}{0.4\textwidth}
		\centering
		\includegraphics[width=\linewidth]{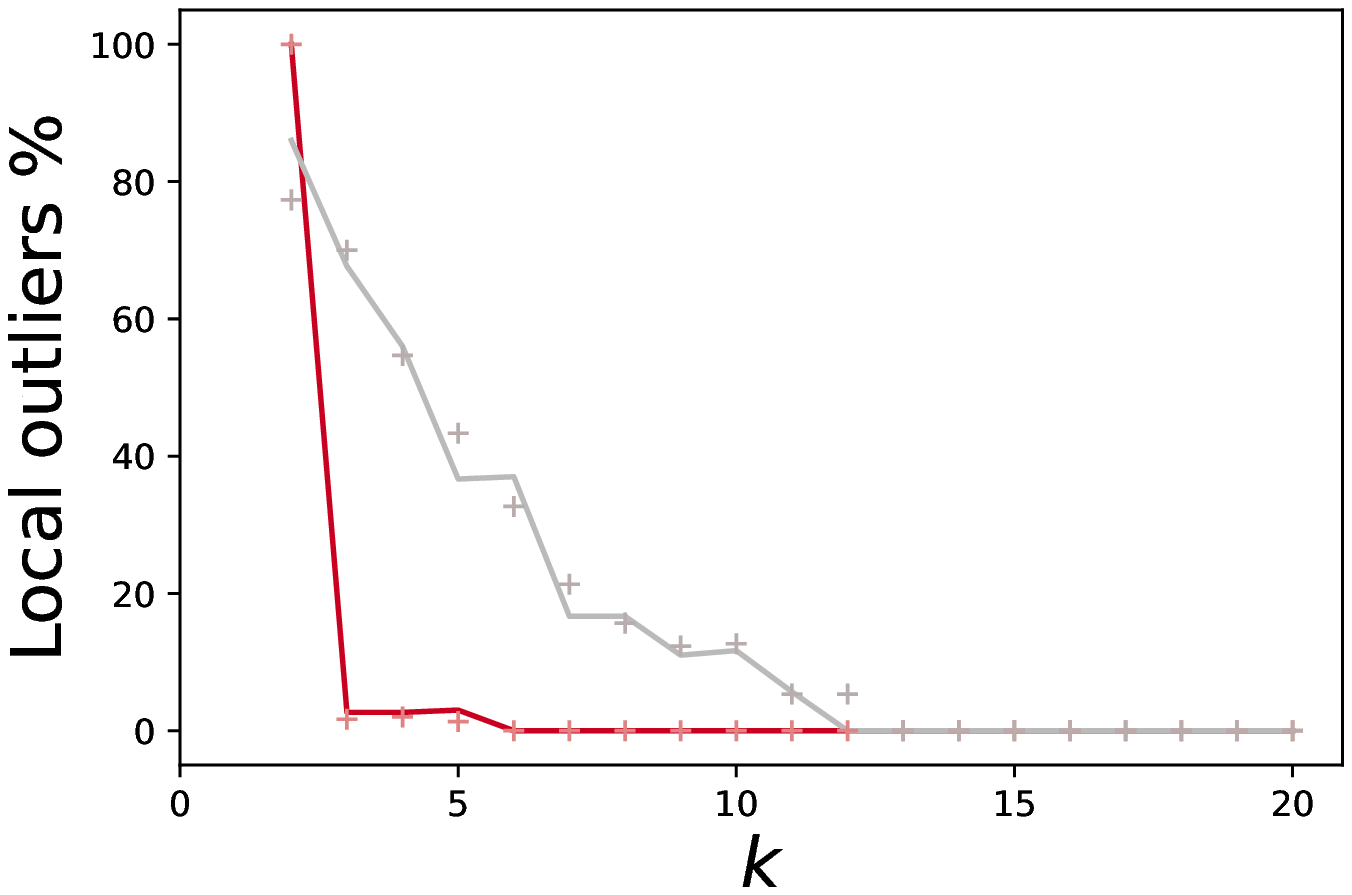}
		\caption{Proximity scores}
		\label{fig:proximity_adv}
	\end{subfigure}
	\begin{subfigure}{0.4\textwidth}
		\centering
		\includegraphics[width=\linewidth]{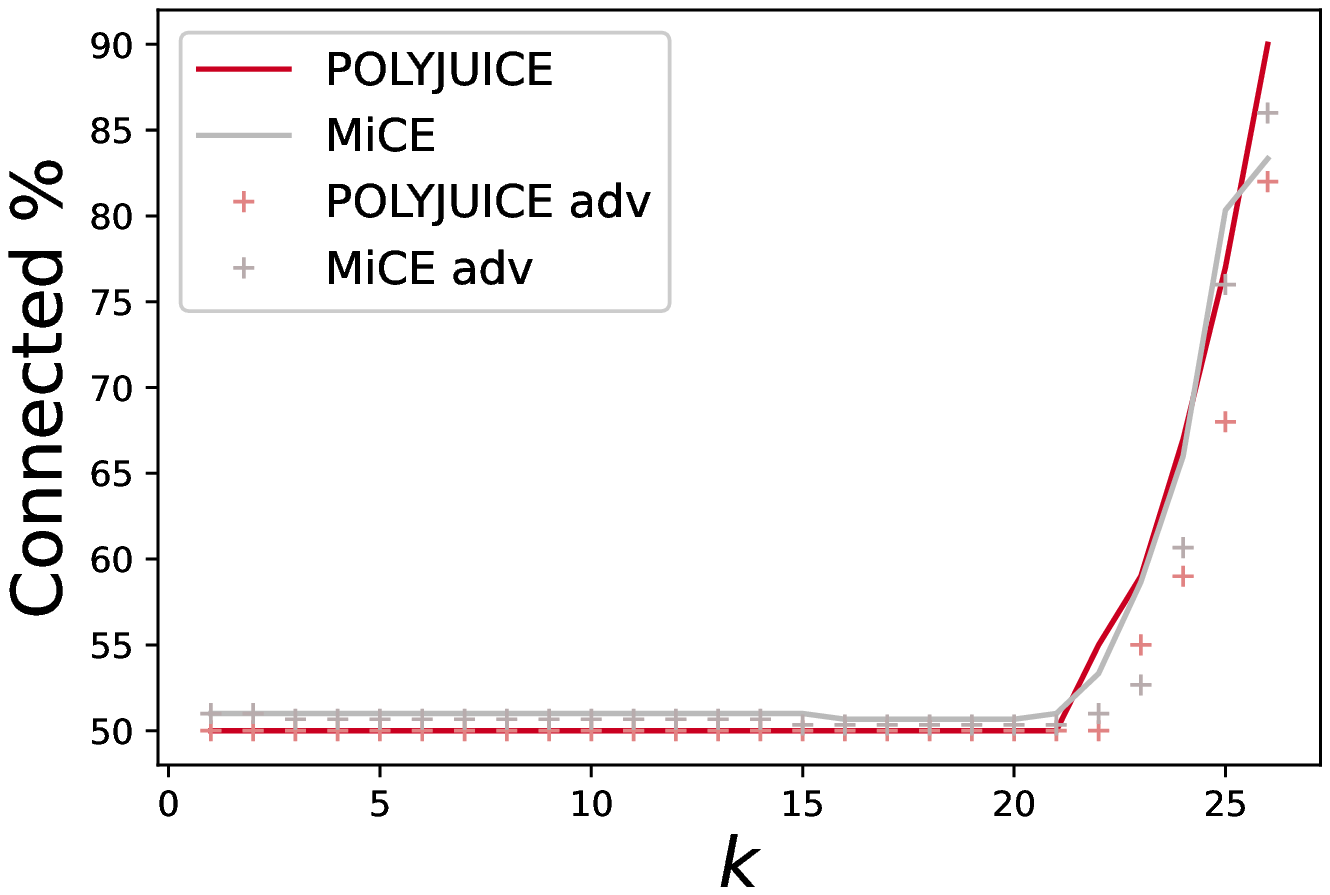}
		\caption{Connectedness scores}\label{fig:connected_adv}
	\end{subfigure}
	\caption{Results with adversarial attacks}
\end{figure}

We generate adversarial perturbations based on semantic similarity \cite{morris2020textattack} on the restaurant reviews. The adversarial inputs are then fed into POLYJUICE and MiCE for a counterfactual generation. Figure~\ref{fig:proximity_adv} demonstrates that the perturbation had no impact on the proximity behavior of POLYJUICE. Markedly, MiCE's counterfactuals became less in-distribution with ground truth data showing questionable robustness to adversarial attacks. The connectedness scores are not affected for both methods as shown in Figure~\ref{fig:connected_adv}. 

\begin{figure}[h]
	\centering
	\begin{subfigure}{0.25\textwidth}
		\centering
		\includegraphics[width=0.95\linewidth]{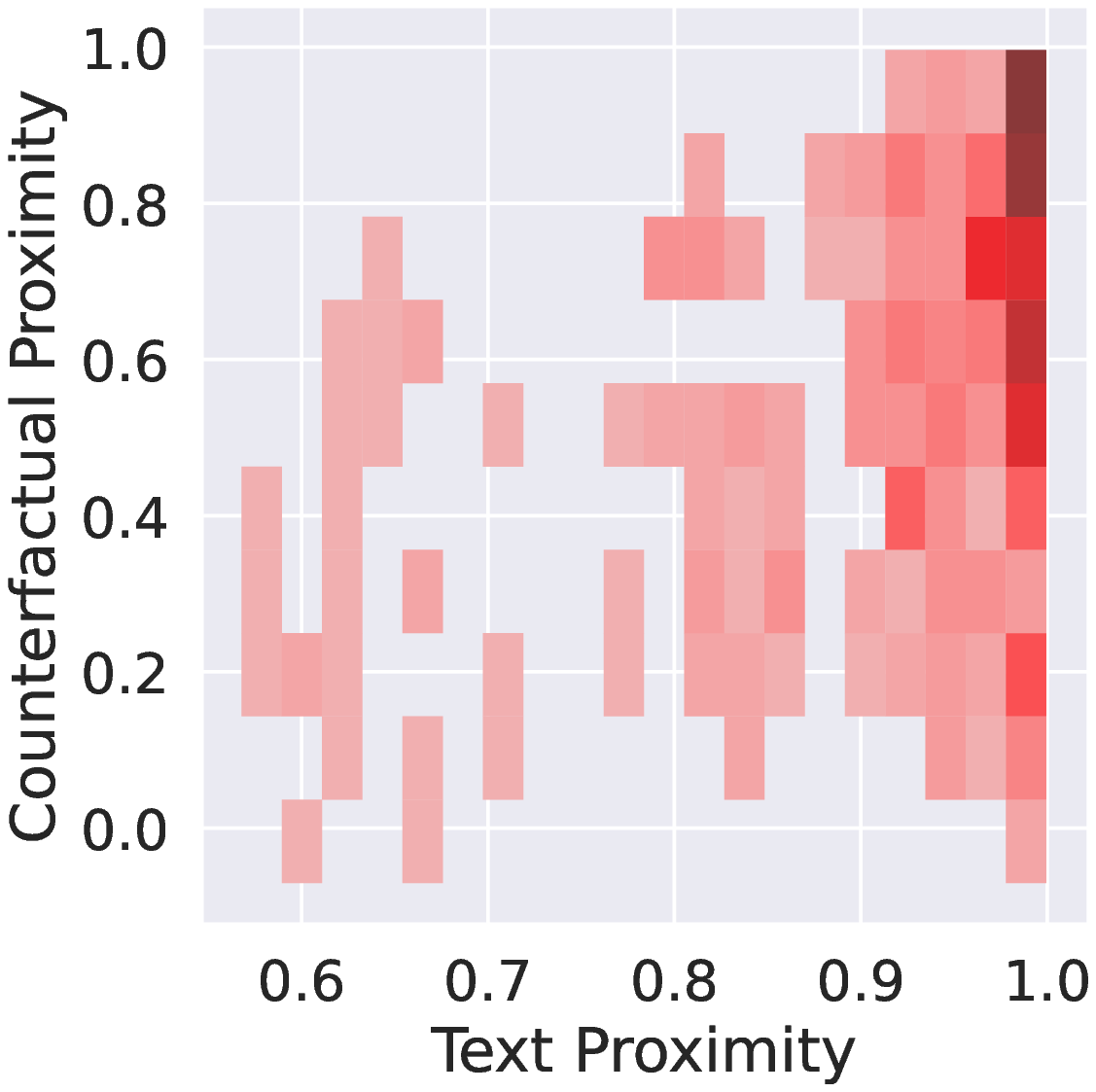}
		\caption{POLYJUICE}
	\end{subfigure}%
	\begin{subfigure}{0.25\textwidth}
		\centering
		\includegraphics[width=0.95\linewidth]{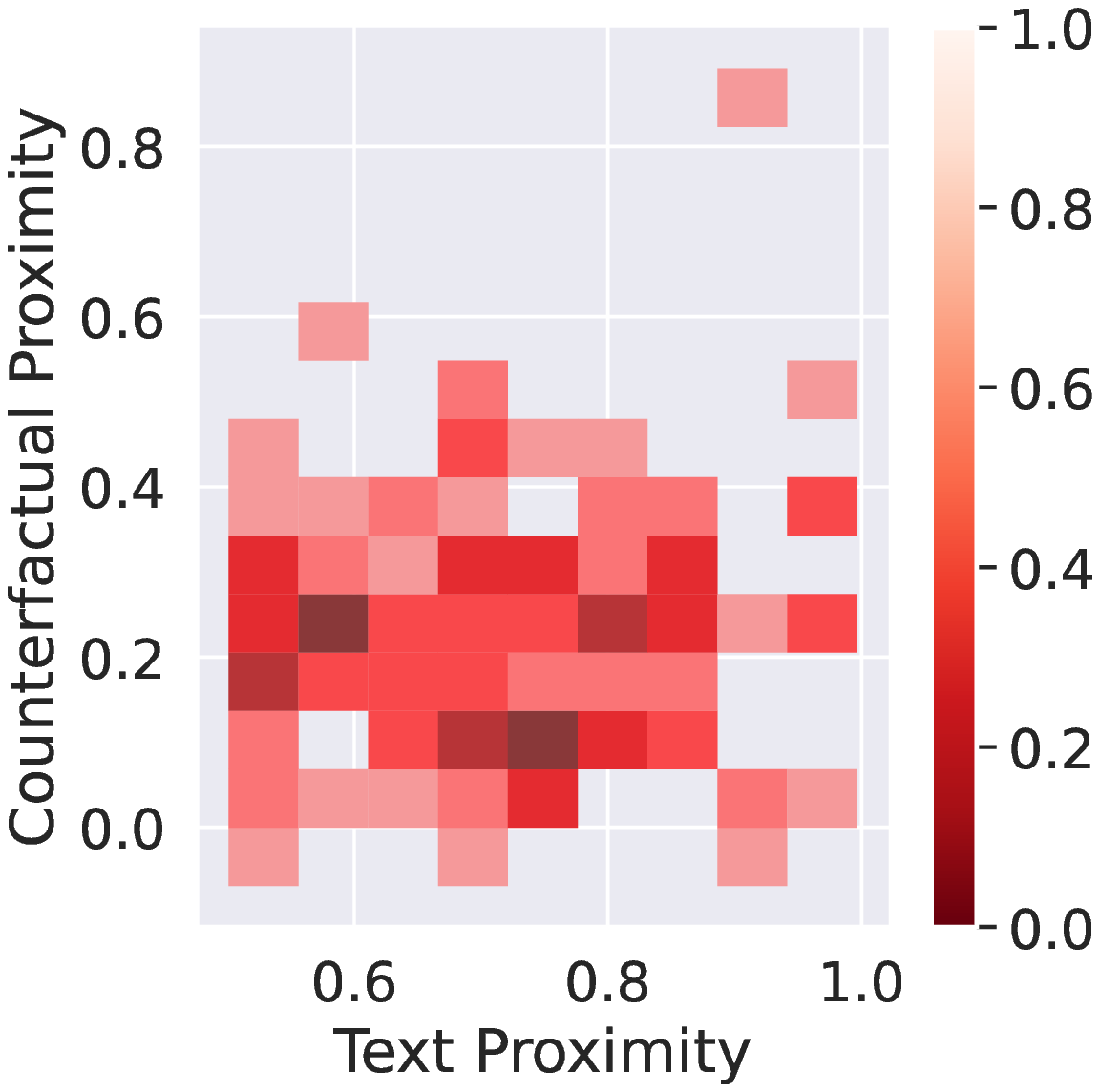}
		\caption{MiCE}
	\end{subfigure}
	\caption{Distribution of the cosine similarity of the generated counterfactuals with adversarial attacks}
	\label{fig:prox_adv}
\end{figure}

Finally, we visualize how the generated counterfactuals are affected when inputs are perturbed. Figure~\ref{fig:prox_adv} shows the distribution of the cosine similarities between $x_{\text{cf}}$ (the counterfactual of the original input, $x$) and $x_{\text{cf}}^{\text{adv}}$ (the counterfactual of its adversarial counterpart, $x^{\text{adv}}$) with respect to the similarity between $x$ and $x^{\text{adv}}$ on a sample of 300 points. POLYJUICE scores higher similarities between counterfactuals showing more robustness to adversarial attacks. Since POLYJUICE does not rely on gradient descent to reach recourse, its results are per the discussion of \cite{slack2021counterfactual} on the problematic behavior of gradient-based counterfactual search on robustness. 

While we are aware of the wide range of adversarial textual attacks, we restrict our experiment to semantic similarity and leave the rest for future inspection. We underline that this experiment is different from the attacks discussed in Section~\ref{sec:adversarial}.
Rather than attacking the classifier, we perturb its input and feed it into a counterfactual method to study whether the latter is robust to adversarial attacks.

\subsection{Comparison to Existing Metrics}
We compute existing evaluation metrics, BLEU and Self-BERT mainly, on the generated counterfactuals. On average, POLYJUICE counterfactuals achieve a BLEU score of 0.38 as opposed to a 0.32 score achieved by MiCE. Self-BERT scores were higher, where POLYJUICE and MiCE achieve 0.95 and 0.92 scores respectively. 

\begin{figure}[h]
	\centering
	\begin{subfigure}{0.25\textwidth}
		\centering
		\includegraphics[width=0.95\linewidth]{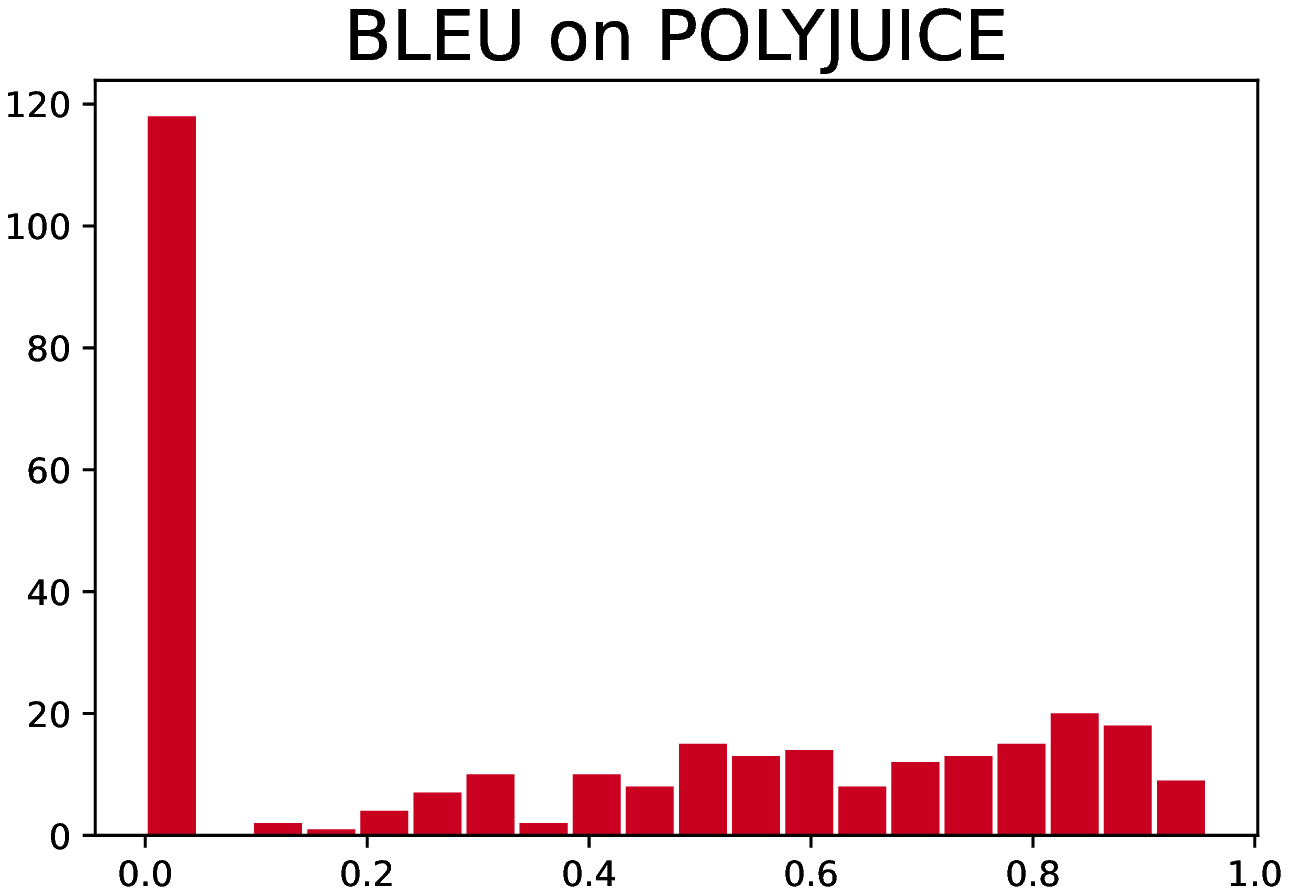}
	\end{subfigure}%
	\begin{subfigure}{0.25\textwidth}
		\centering
		\includegraphics[width=0.95\linewidth]{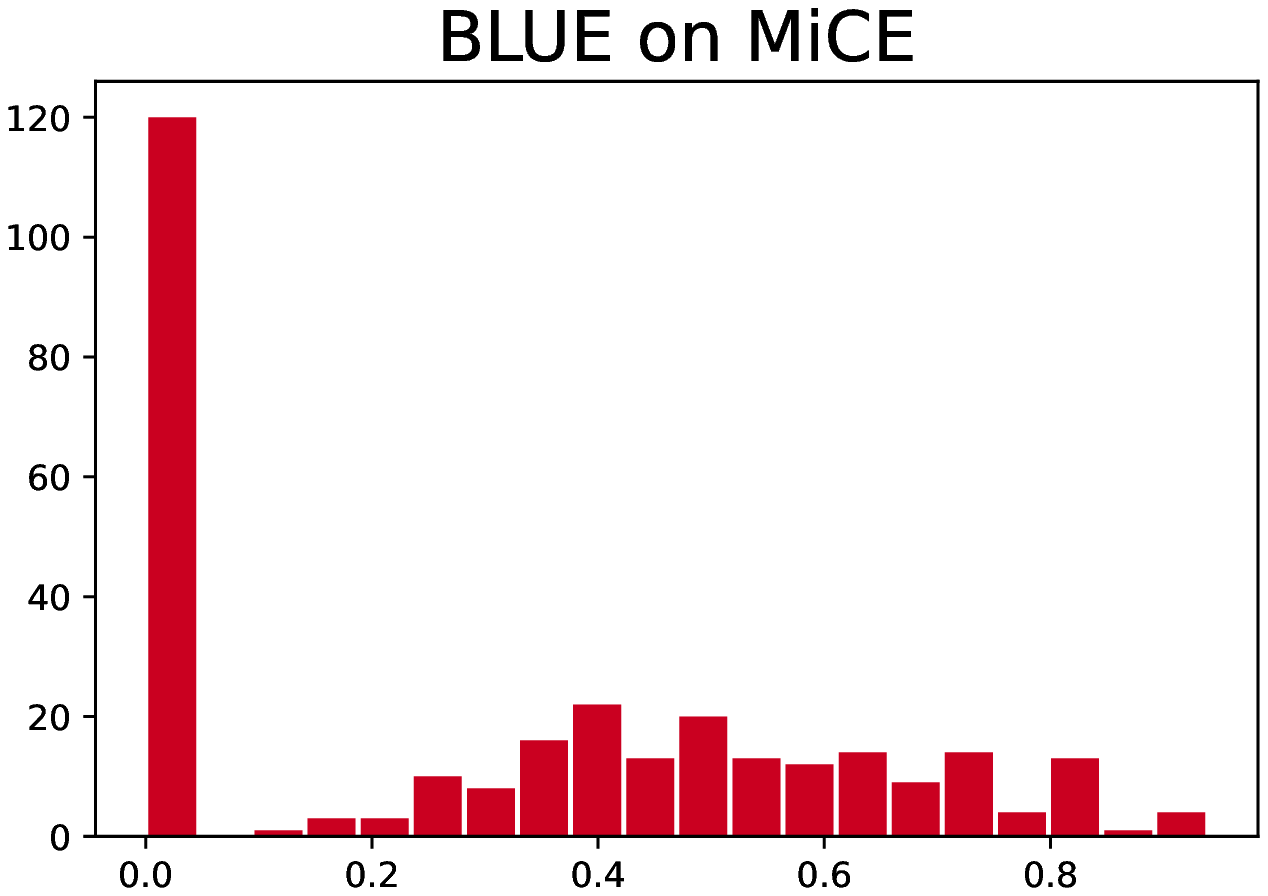}
	\end{subfigure}
 \begin{subfigure}{0.25\textwidth}
		\centering
		\includegraphics[width=0.95\linewidth]{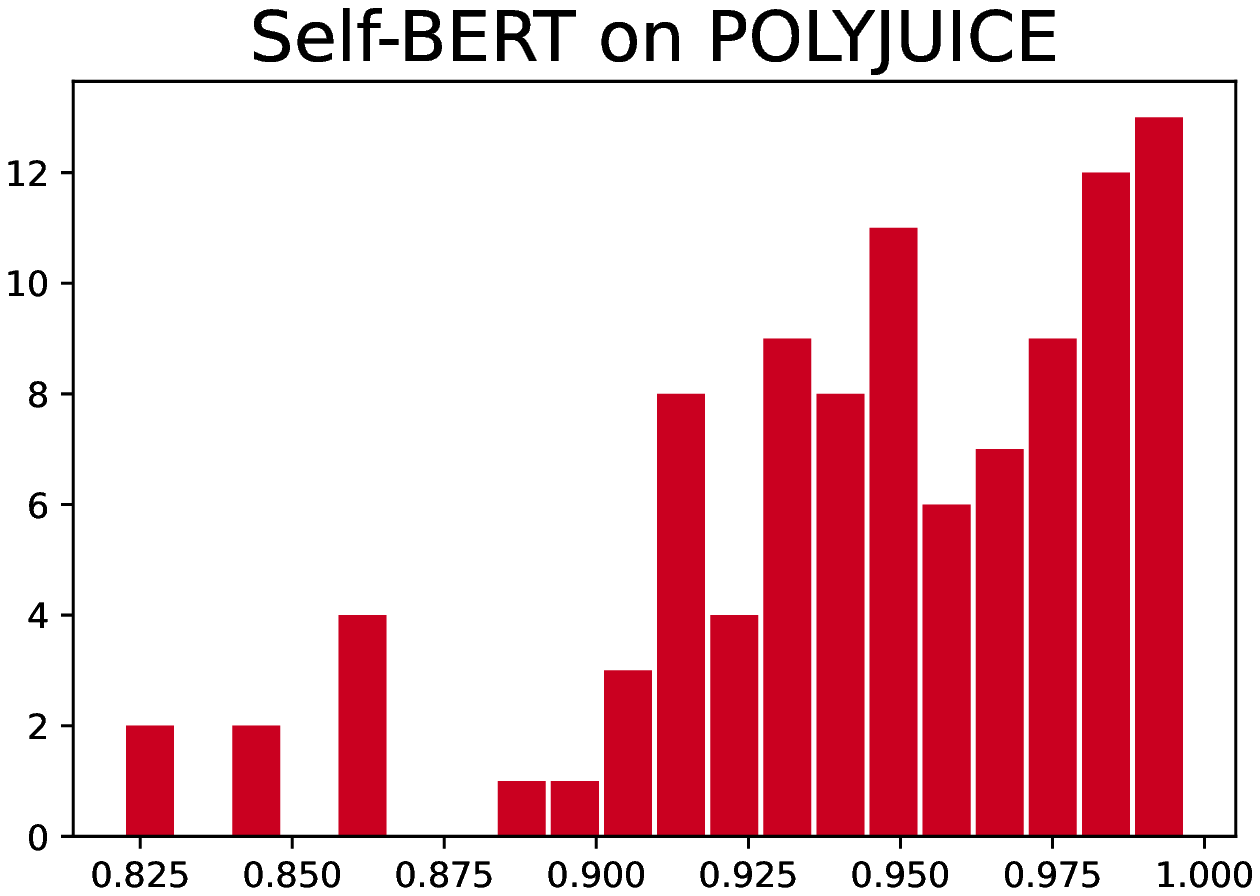}
	\end{subfigure}%
	\begin{subfigure}{0.25\textwidth}
		\centering
		\includegraphics[width=0.95\linewidth]{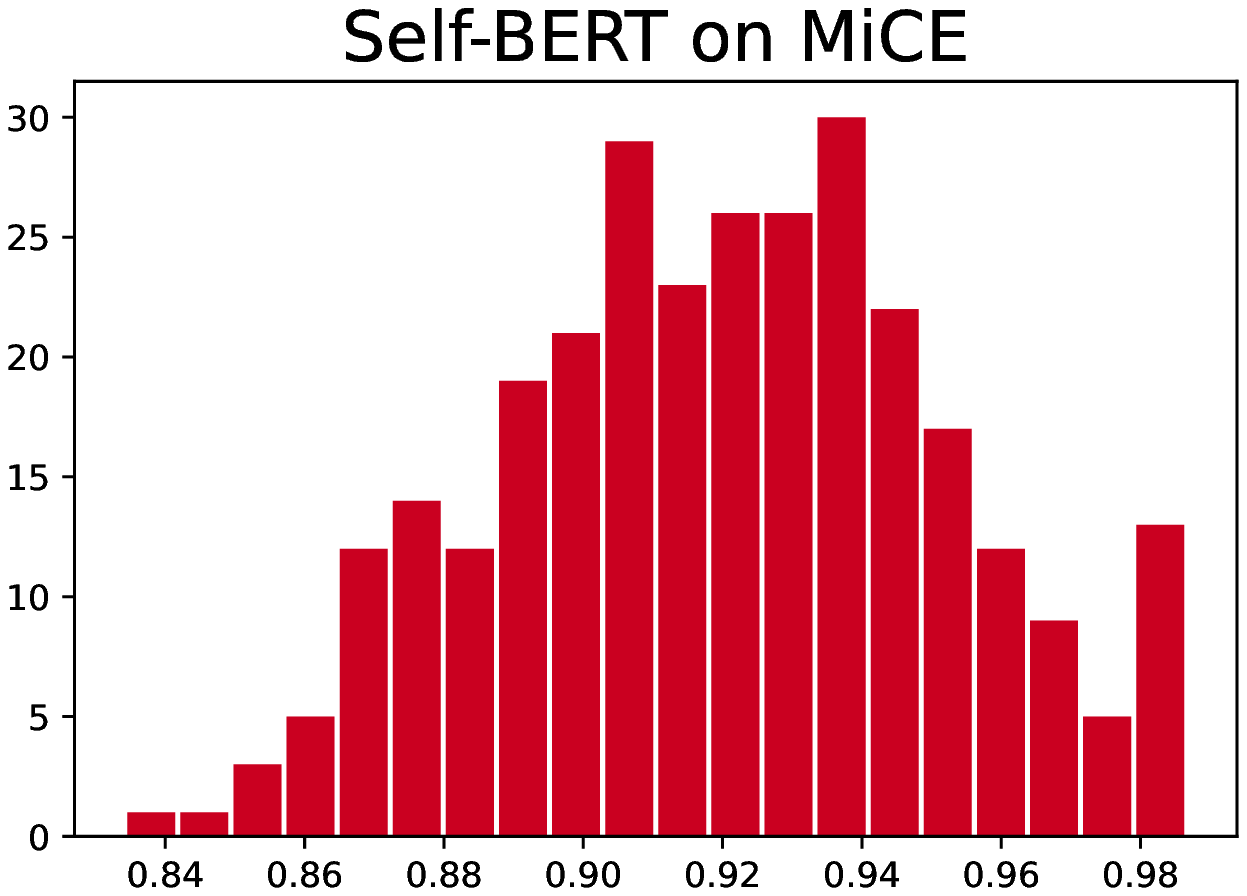}
	\end{subfigure}
	\caption{Distribution of the BLEU and Self-BERT scores}\label{fig:existing_metrics}
\end{figure}

The results show a slight improvement of POLYJUICE over MiCE which confirms our findings highlighting again the importance of latent representations. Figure~\ref{fig:existing_metrics} shows the distribution of the scores on the counterfactuals generated by POLYJUICE and MiCE.

\subsection{Discussion}
The fundamental difference between POLYJUICE and MiCE can be traced to word representations. The former anticipates latent space encodings while the latter operates at the textual level. Hence, we will interpret their faithfulness through the word representation lens. 

Proximity results were not consistent. Higher $P(\bm{x}_{cf})$ scores are reported with MiCE while lower outlier factors are observed with POLYJUICE. One can thus say, that relative to $d(\bm{x}, \bm{x}_{cf})$ edits on the textual level achieve higher proximity. Considering a cluster of ground truth inputs with the same class as the counterfactual, POLYJUICE is shown to obey the input distribution in generating contrastive texts. We also call attention to the fluency filtering layer of POLYJUICE which yields better reachability. These results hint at the connection between latent representations and the attainability of generated counterfactuals.

Conversely, connectedness scores do not show any substantial difference. 
POLYJUICE is shown to be more locally stable and more robust to adversarial attacks. The results make intuitive sense as the distances are computed based on latent representations that are used by POLYJUICE in their contrastive search. Hence, latent representation of words (instead of textual ones) can serve the algorithmic stability of recourse methods. 


\section{Conclusion}\label{sec:conc}
Counterfactual methods go beyond interpretability and offer practical explanations that comply with the social and algorithmic aspects of explainability. In this work we chart the path towards contrastive methods in NLP with a common evaluation scheme inspired by \textit{faithfulness}. We present the limitations of traditional explainability which are leveraged with recourse methods, in NLP mainly. 

We further define \textit{faithfulness} of textual explanations and present corresponding computation schemes. Our benchmarks on two famous methods, POLYJUICE and MiCE, show that better algorithmic stability and attainability are achieved in the former, highlighting the importance of latent representation in the counterfactual search strategy. We highlight the vulnerabilities of textual recourse methods against semantic adversarial attacks. Three immediate steps in this line of work are the mitigation of the ``unconnected'' counterfactuals by posing connectedness constraints on the search strategy, the enhancement of the stability when textual edits are employed, and the investigation of textual attacks on recourse methods.


\section{Limitations}
While our work addresses one of the limitations of counterfactual textual explanations, \textit{faithfulness} evaluation specifically, it has its own restrictions. 
First, the connectedness aspect of \textit{faithfulness} is computed based on neighbors sampled from a validation set. This evaluation reflects the plausibility of generated counterfactuals based on the data at hand. A more generalizable plausibility requires sufficiently large validation sets. 

One can also argue that \textit{faithfulness} heavily depends on the distance notion which is based on transformers' encodings. Although transformers are state-of-the-art language models that successfully encode syntax and semantics, their performance is crucial for a \textit{faithful} evaluation. Extension to other languages requires careful consideration. This is mainly due to the counterfactual generation process operating differently with different morphologies and so does the distance measure. 

Finally, the intriguing relation between counterfactuals and AEs can motivate the use of the former to improve models' robustness against AEs. Informing practitioners of their potential harm is a key responsibility to preventing unfavorable manipulations.  


\bibliographystyle{acl_natbib}
\bibliography{refs}
\end{document}